 \documentclass[pmlr,twocolumn,10pt]{jmlr} 

\usepackage{booktabs}
\usepackage{siunitx}

\usepackage[switch]{lineno}

\usepackage{booktabs}
\usepackage{makecell}
\usepackage{makecell}
\usepackage{adjustbox}
\usepackage{pifont}
\usepackage{multirow}
\usepackage{float}
\usepackage{listings}
\usepackage{xcolor} 
\usepackage{enumitem}

\newcommand{\cmark}{\ding{51}}
\newcommand{\xmark}{\ding{55}}

\newcommand{\equal}[1]{{\hypersetup{linkcolor=black}\thanks{#1}}}

\theorembodyfont{\upshape}
\theoremheaderfont{\scshape}
\theorempostheader{:}
\theoremsep{\newline}

\jmlrvolume{297}
\jmlryear{2025}
\jmlrsubmitted{LEAVE UNSET}
\jmlrpublished{LEAVE UNSET}
\jmlrworkshop{Machine Learning for Health (ML4H) 2025} 

\title[FHIR-AgentBench]{FHIR-AgentBench: Benchmarking LLM Agents for Realistic Interoperable EHR Question Answering}

\author{%
\Name{Gyubok Lee}\equal{These authors contributed equally}\footnotemark[2] \Email{gyubok.lee@kaist.ac.kr}\\
\Name{Elea Bach}\footnotemark[1]\footnotemark[3] \Email{eleabach@verily.com}\\
\Name{Eric Yang}\footnotemark[3] \Email{eryang@verily.com}\\
\Name{Tom Pollard}\footnotemark[4] \Email{tpollard@mit.edu}\\
\Name{Alistair Johnson}\footnotemark[4] \Email{aewj@mit.edu}\\
\Name{Edward Choi}\footnotemark[2] \Email{edwardchoi.kaist.ac.kr}\\
\Name{Yugang Jia}\footnotemark[3] \Email{yugang@verily.com}\\
\Name{Jong Ha Lee}\footnotemark[3] \Email{jonghalee@verily.com}\\
\addr \footnotemark[2] Korea Advanced Institute of Science \& Technology, South Korea \\
\addr \footnotemark[3] Verily Life Sciences, USA \\
\addr \footnotemark[4] Massachusetts Institute of Technology, USA \\
\addr \footnotemark[1] Co-first authors \\
}

\begin{document}

\maketitle

\begin{abstract}
The recent shift toward the Health Level Seven Fast Healthcare Interoperability Resources (HL7 FHIR) standard opens a new frontier for clinical AI, demanding LLM agents to navigate complex, resource-based data models instead of conventional structured health data. However, existing benchmarks have lagged behind this transition, lacking the realism needed to evaluate recent LLMs on interoperable clinical data. To bridge this gap, we introduce FHIR-AgentBench—a benchmark that grounds 2,931 real-world clinical questions in the HL7 FHIR standard.
Using this benchmark, we systematically evaluate agentic frameworks, comparing different data retrieval strategies (direct FHIR API calls vs. specialized tools), interaction patterns (single-turn vs. multi-turn), and reasoning strategies (natural language vs. code generation).
Our experiments highlight the practical challenges of retrieving data from intricate FHIR resources and the difficulty of reasoning over them—both of which critically affect question answering performance.
\end{abstract}

\paragraph*{Data and Code Availability}
Data and code are publicly available on our GitHub repository     \href{https://github.com/glee4810/FHIR-AgentBench/tree/master}{FHIR-AgentBench}.

\paragraph*{Institutional Review Board (IRB)}
IRB approval is not required for this work. The patient records used in this work are from the PhysioNet website and licensed under the Open Data Commons Open Database License v1.0\footnote{\url{https://opendatacommons.org/licenses/odbl/1-0/}}.

\begin{table*}[t!]
\centering
\footnotesize
\renewcommand{\arraystretch}{1.2}
\caption{Comparison of FHIR-AgentBench with existing EHR QA benchmarks. The use of Synthea~\citep{walonoski2018synthea} means that questions are grounded in synthetic patient data. Real Q. denotes whether the questions reflect actual clinician needs. $\triangle$ denotes questions collected by shadowing clinicians in an ICU setting.}
\vspace{-2mm}
\begin{adjustbox}{width=\linewidth}
\label{tab:dataset_stats}
\begin{tabular}{lccccc}
\hline
\textbf{Benchmark}
& \textbf{Source Data}
& \textbf{\# Question}
& \textbf{Real Q.}
& \textbf{FHIR}
& \textbf{QA-Specific}
\\
\hline
MIMICSQL~\citep{wang2020textsql}
& MIMIC-III
& 10,000
& \xmark
& \xmark
& \cmark
\\
$\text{FHIR}_{\text{DATA}}$~\citep{soni2020using}
& Synthea
& 966
& \cmark
& \cmark
& \cmark
\\
EHRSQL~\citep{lee2023ehrsql}
& eICU \& MIMIC-III
& 24,411
& \cmark
& \xmark
& \cmark 
\\
$\text{ICU}_{\text{DATA}}$~\citep{soni2023quehry}
& Synthea
& 400
& $\triangle$
& \xmark
& \cmark
\\
EHRSQL-2024~\citep{lee-etal-2024-overview}
& MIMIC-IV
& 7,454
& \cmark
& \xmark
& \cmark
\\
\citet{kothari2025questionansweringpatientmedical}
& Synthea
& 5,000
& \xmark
& \cmark
& \cmark 
\\
MedAgentBench~\citep{jiang2025medagentbenchrealisticvirtualehr}
& STARR
& 300
& \cmark
& \cmark
& \xmark 
\\
FHIR-AgentBench (Ours) 
& MIMIC-IV-FHIR
& 2,931
& \cmark
& \cmark
& \cmark 
\\
\hline
\end{tabular}
\end{adjustbox}
\end{table*}

\section{Introduction}
\label{sec:intro}

Large Language Model (LLM) agents in healthcare applications are rapidly being developed to reduce administrative burdens, support diagnostic reasoning, and personalize patient care~\citep{thirunavukarasu2023large, nazi2024large, xiaolan2025evaluating}. However, for these agents to move from research concepts to deployable tools, they require integration with the data ecosystem of healthcare systems. Increasingly, these systems are adopting the Health Level Seven Fast Healthcare Interoperability Resources (HL7 FHIR) standard~\citep{hl7_fhir_2025} to organize, share, and query electronic health data in a codified, shared schema---creating an important opportunity to develop models that can operate across different health systems. This transition expands the challenge from interacting with electronic health record (EHR) databases through query languages such as SQL to reasoning across interoperable health data resources~\citep{soni2023quehry, jiang2025medagentbenchrealisticvirtualehr}. 

In particular, the ability of an agent to navigate the complex, graph-like structure inherent in FHIR resources, retrieve relevant information, and synthesize an answer is crucial. Unlike traditional relational databases composed of flat tables, FHIR organizes data as discrete, nested resources interconnected by references. As a result, agents must effectively traverse this graph structure, frequently requiring multiple retrieval steps to gather and synthesize the information needed to answer a clinical question. This capability is essential for enabling question answering (QA) systems to reduce the friction of data access for clinicians and researchers.

However, benchmarks for evaluating such agents have lagged behind interoperability trends like FHIR and lack realistic settings to assess the capabilities of recent LLMs.
Existing benchmarks for question answering over structured EHRs, such as MIMICSQL~\citep{wang2020textsql} and EHRSQL~\citep{lee2023ehrsql}, primarily evaluate an agent's ability to translate natural language questions into SQL queries operating on non-standard, healthcare system–specific data schemas.
In contrast, benchmarks addressing interoperability have either relied on synthetic data~\citep{soni2020using, kothari2025questionansweringpatientmedical}, which fall short of capturing the complexity of real patient records, or focused on broad tasks~\citep{jiang2025medagentbenchrealisticvirtualehr}, lacking a targeted evaluation of question answering that requires extensive patient record retrieval and reasoning (see Table~\ref{tab:dataset_stats} for comparison).

\begin{figure*}[t!]
\centering
\includegraphics[width=\linewidth]{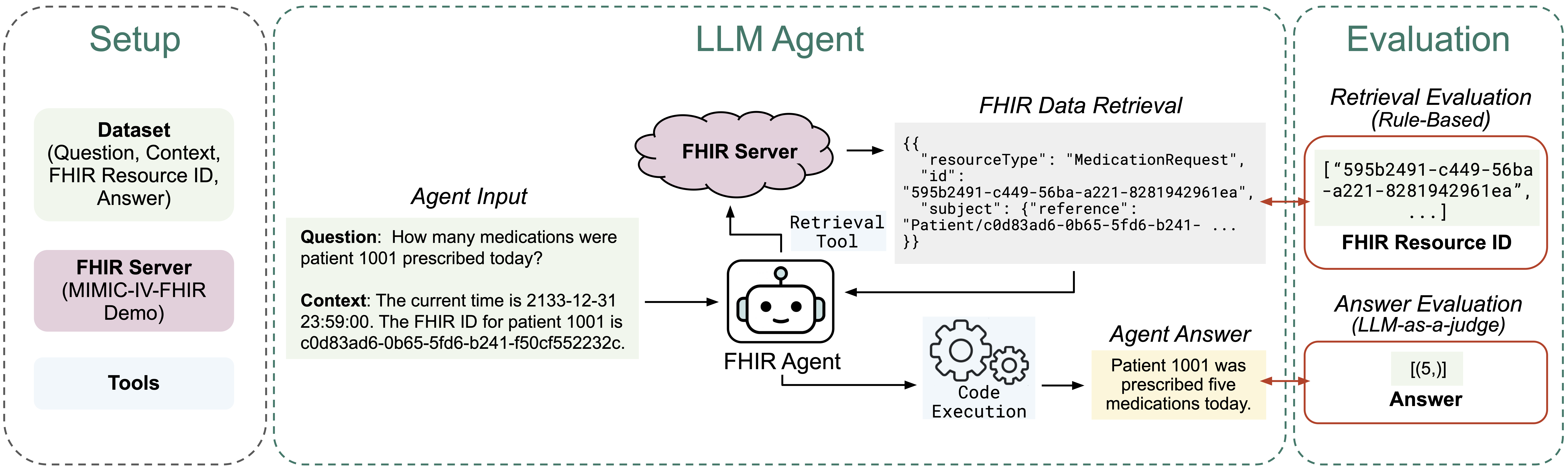}
\vspace{-5mm}
\caption{Overview of FHIR-AgentBench.}
\label{fig:overall_benchmark}
\end{figure*}

To address this gap, we introduce \textbf{FHIR-AgentBench}, a benchmark that integrates three critical dimensions for realistic EHR question answering: \emph{clinical question authenticity}, \emph{patient-record complexity}, and \emph{FHIR system fidelity}. FHIR-AgentBench combines clinician-sourced questions from EHRSQL with answers derived from de-identified MIMIC-IV-FHIR patient records and their associated FHIR resources (analogous to data rows in databases).
Our benchmark challenges agents with real-world clinical QA tasks such as temporal reasoning, navigating thousands of records, handling case-sensitive terminologies, and addressing queries with no relevant results. By evaluating a range of agentic frameworks—from naive LLMs to tool- and code-augmented single- and multi-turn agents—we measure both retrieval accuracy and answer correctness, offering insights for future agent design (see Figure\ref{fig:overall_benchmark} for the overall benchmark pipeline).

The main contributions of our work are summarized as follows:

\begin{itemize}
    \item \textbf{A realism-grounded FHIR QA benchmark.} We introduce FHIR-AgentBench, which grounds 2,931 clinician-sourced questions in MIMIC-IV-FHIR, enhancing question authenticity, data complexity, and interoperability fidelity in agent evaluation for EHR QA.
    \item \textbf{Comprehensive agent evaluation.} We systematically compare naive LLMs with tool- and code-augmented agents in both single- and multi-turn settings. We measure answer correctness and retrieval quality (resource-level precision/recall) to identify performance bottlenecks and provide actionable insights.
    \item \textbf{Transparent reproducibility.} We publicly release the dataset, SQL-to-FHIR conversion pipeline, agent implementations, and evaluation suite to enable transparent, reproducible agent research for EHR QA.
\end{itemize}

\section{Related Works}
\label{sec:related_work}

\paragraph{SQL-based EHR QA benchmarks}

Early benchmarks for EHR question answering primarily focus on translating natural language queries into SQL over non-standard schemas. 
They vary in question authenticity yet lack fidelity to standardized protocols like HL7 FHIR. 
MIMICSQL~\citep{wang2020textsql} employs synthetic questions on MIMIC-III~\citep{pollard2018eicu}, while EHRSQL~\citep{lee2023ehrsql} and EHRSQL-2024~\citep{lee-etal-2024-overview} increase realism by curating clinician-sourced questions from over 200 healthcare professionals, grounded in eICU, MIMIC-III~\citep{johnson2019mimiciii}, and MIMIC-IV~\citep{johnson2023mimiciv} databases. 
These benchmarks capture authentic clinical needs but remain tied to SQL and site-specific schemas, which limits interoperability and real-time exchange.
In contrast, our benchmark combines real clinical questions and real patient records from EHRSQL-2024 (grounded in MIMIC-IV) with the system fidelity of MIMIC-IV-FHIR~\citep{bennett2024mimicivfhir} to address existing interoperability limitations.

\paragraph{FHIR-integrated QA benchmarks}

Several benchmarks address the shift toward FHIR for question answering. \citet{soni2020using} and \citet{kothari2025questionansweringpatientmedical} evaluate models using clinician-sourced questions but rely on Synthea~\citep{walonoski2018synthea}, a synthetic patient dataset. Although these works integrate FHIR, they overlook real-world challenges such as handling thousands of medical events and managing missing or unnormalized clinical terms. Our benchmark targets these gaps by grounding clinician-sourced questions in de-identified, real patient records from MIMIC-IV-FHIR.

\paragraph{Agent-based clinical benchmarks and evaluation}

Recent advances in clinical AI highlight the emergence of agentic workflows in LLMs, allowing models to plan, call tools, and reason across multiple steps to interact with clinical data.
MedAgentBench~\citep{jiang2025medagentbenchrealisticvirtualehr} conducts a broad assessment of agents' clinical competencies in a virtual EHR, supporting multi-turn dialogs across 10 task categories that range from information retrieval to simulated clinical actions such as order placement using FHIR on a small scale of patient data\footnote{While MedAgentBench covers diverse clinical workflows, FHIR-AgentBench focuses on testing an agent's ability to handle complex multi-step questions that require deep reasoning over real-world patient histories (often involving hundreds or thousands of FHIR resources).}. 
MedAgentGym~\citep{xu2025medagentgymtrainingllmagents} provides a training environment that scales code-centric medical reasoning over clinical notes, laboratory reports, structured EHR tables, and biological sequences, exposing agents to programmatic tool use and iterative planning with explicit feedback loops.
ArchEHR-QA~\citep{soni2025archehrqa} explores multi-agent pipelines for clinical notes, combining ReAct-style reasoning with tool-based retrieval and API calls for document understanding and evidence gathering. Compared with these efforts, our work offers a focused, FHIR-specific benchmark for complex question answering over real MIMIC-IV-FHIR records and reports controlled ablations across interaction pattern (single-turn vs. multi-turn), reasoning strategy (natural language vs. code), and retrieval mechanism (direct FHIR API calls vs. specialized tools), along with answer correctness and resource-level retrieval metrics.

\begin{figure}[t!]
\centering
\includegraphics[width=\linewidth]{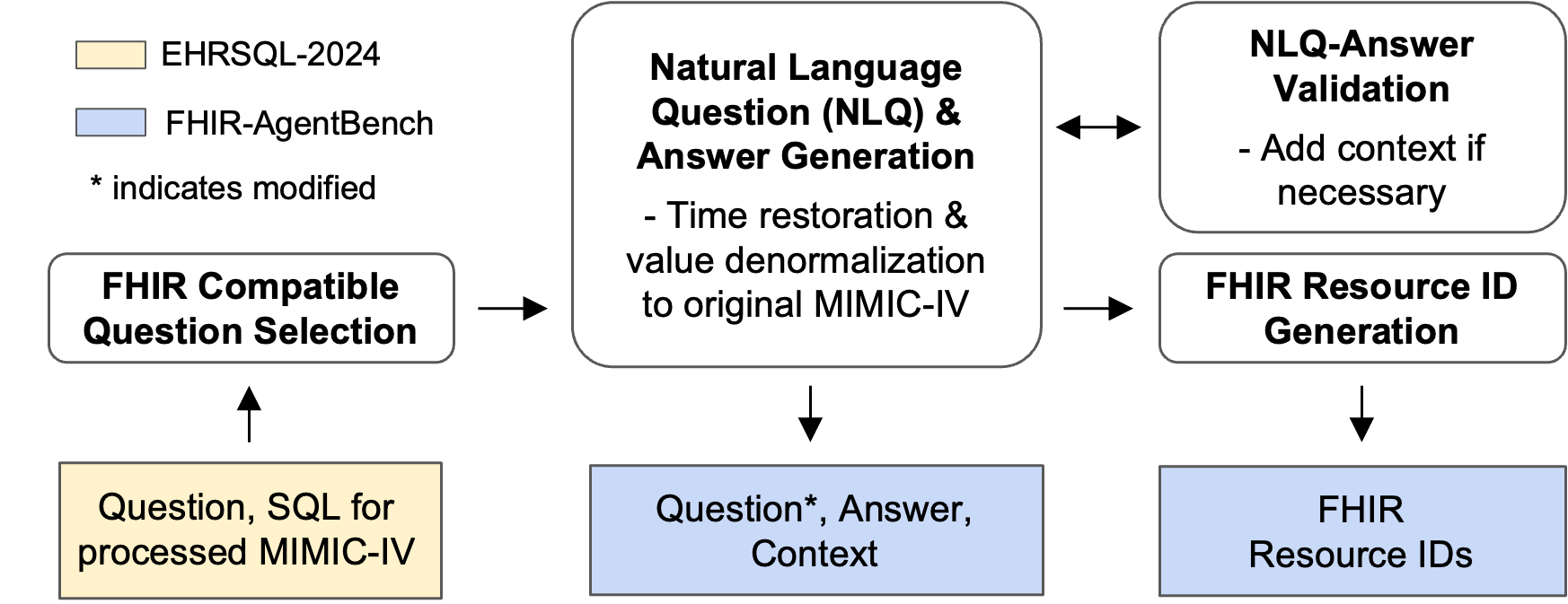}
\vspace{-5mm}
\caption{Overview of the construction process for FHIR-AgentBench.}
\label{fig:benchmark_construction}
\end{figure}

\section{Benchmark Construction}
\label{sec:benchmark}

In this section, we describe the benchmark construction process, which involves (1) selecting FHIR-compatible questions, (2) generating natural language questions (NLQs) and answers, (3) verifying the NLQ–answer pairs, and (4) generating FHIR resource IDs that specify the patient data required to answer each question. The overall construction process is illustrated in Figure~\ref{fig:benchmark_construction}. The FHIR version used in FHIR-AgentBench is L7 FHIR R4 (v4.0.1), which follow the same FHIR version used in MIMIC-IV-FHIR-Demo ~\citep{bennett2024mimicivfhir}.

\begin{table*}[t!]
\centering
\small
\renewcommand{\arraystretch}{1.2}
\caption{Sample data in FHIR-AgentBench.
An answer of 1 indicates ``yes'' (based on SQL results).}
\begin{adjustbox}{width=\linewidth,center}
\begin{tabular}{p{0.32\textwidth} p{0.2\textwidth} p{0.3\textwidth} p{0.2\textwidth}}
\toprule
\textbf{Natural Language Question} & \textbf{Context} & \textbf{FHIR Resource IDs} & \textbf{Answer} \\
\hline
When was the first time the respiratory rate of patient 10018081 was measured to be less than 23.0 today?
& Assume the current time is 2133-12-31 23:59:00.
& \{`Observation': [`b2c4828b-2de2-5116-9485-9e34d05bee40']\}
& [[`2133-12-31 02:00:00']] \\
\hline
When has patient 10018845 been given phosphate test for the last time? 
& When searching for values in the database, account for all variations in letter case and surrounding whitespace. 
& \{`Observation': [`a8407c1c-7737-5d0a-9209-168158821204']\}
& [[`2184-10-10 06:30:00']] \\
\hline
Was patient 10018328 diagnosed with anything in the first hospital visit?
& N/A
& \{`Condition': [`54305768-49b9- 59 8c-8b3b-483b2ec86100', `bcc812bc-a287-5a91-8857-8d0c5dd93fbe', ...\}
& [[1]] \\
\bottomrule
\end{tabular}
\end{adjustbox}
\label{tab:sample_questions}
\end{table*}

\subsection{FHIR Compatible Question Selection}
\label{sec:benchmark_stage1}

To meet realism in questions, data, and system fidelity, we use two sources: (1) EHRSQL-2024~\citep{lee-etal-2024-overview}, which provides question-SQL pairs derived from the processed version of MIMIC-IV-Demo, and (2) the MIMIC-IV-FHIR-Demo dataset~\citep{bennett2024mimicivfhir}\footnote{The demo version supports both commercial and open-source models. Our benchmark is also compatible with the full MIMIC-IV-FHIR.}, which offers real, de-identified patient data in the mandated FHIR format. 

From EHRSQL, we select the subset of questions that focus on a single patient, as these are most representative of point-of-care clinical inquiries. We exclude multi-patient questions (e.g., ``How many patients in their 50s have disease A?'') because the high retrieval latency associated with such queries (e.g., server-side pagination, rate limits, and large payload processing) dominates the overall evaluation time, hindering our core objective of assessing the agent's capabilities. The retained single-patient questions are nonetheless complex, requiring an average of 14.5 FHIR resources to answer, with the most complex cases involving over 2,000 resources. We also exclude questions that require using information unavailable in the FHIR resources.

\subsection{Natural Language Question and Answer Generation}
\label{sec:benchmark_stage2}

A critical challenge in using EHRSQL to link patient data in MIMIC-IV is that EHRSQL is built on a processed version of MIMIC-IV, where data is normalized (e.g., all values lower-cased, timestamps shifted to a fixed year). Thus, ensuring question compatibility with the raw, unnormalized MIMIC-IV-FHIR data requires a meticulous restoration process.

First, we restore the correct anchor years\footnote{anchor year is the de-identified year when the patient was admitted in MIMIC-IV.} for each patient. For questions containing absolute dates (e.g., ``on 2100-10-21''), we update the date to the patient's anchor year. For questions with relative time expressions (e.g., ``in the last 24 hours''), we add explicit context to the question referencing the anchor year (e.g., ``Assume the current time is \{anchor\_year\}-10-21.'').

Next, we reverse the normalization of clinical terms. We map the lower-cased, whitespace-stripped terms in EHRSQL back to their original, varied formats found in the raw MIMIC-IV data (e.g., ``lorazepam'' $\rightarrow$ ``Lorazepam'' or ``LORazepam'').
 
After this restoration, we re-execute the modified SQL queries on the raw MIMIC-IV database to generate the final answers. This process reveals a crucial real-world scenario: approximately 24\% of questions correctly yielded an empty answer. We deliberately retain these cases to challenge the common benchmark assumption that a non-empty answer always exists.

\subsection{Natural Language Question-Answer Validation}
\label{sec:benchmark_stage3}

While generating NLQs and answers, we iteratively review whether the natural language questions truly correspond to the answers based on the modified SQL. If additional context for the questions is required due to assumptions in SQL annotation (e.g., ``SpO2'' is mapped to ``oxygen saturation pulse oximetry'' in MIMIC-IV), we provide it when necessary to ensure accurate evaluation of the agents. This step is particularly important since our benchmark, unlike EHRSQL-2024, does not include a training set.
\vspace{-2mm}

\subsection{FHIR Resource ID Generation}
\label{sec:benchmark_stage4}
To evaluate an agent's ability to retrieve the correct data, we generate a set of all FHIR resource IDs required to answer each question. This process involves: (1) modifying the original SQL queries to return all data rows, not just the answer, required to answer the question; and (2) extracting the unique ID for each FHIR resource corresponding to these rows \citep{bennett2024mimicivfhir}. An overview of this process is provided in \appendixref{fig:translation_example}. When answers are non-empty, the corresponding FHIR IDs are always present. In contrast, when answers are empty, no corresponding FHIR IDs exist. This provides an unambiguous ground truth for evaluating retrieval precision and recall. 
\vspace{-2mm}

\subsection{Dataset Composition and Statistics}

\begin{figure}[t!]
\centering
\vspace{-5mm}
\includegraphics[width=8cm]
{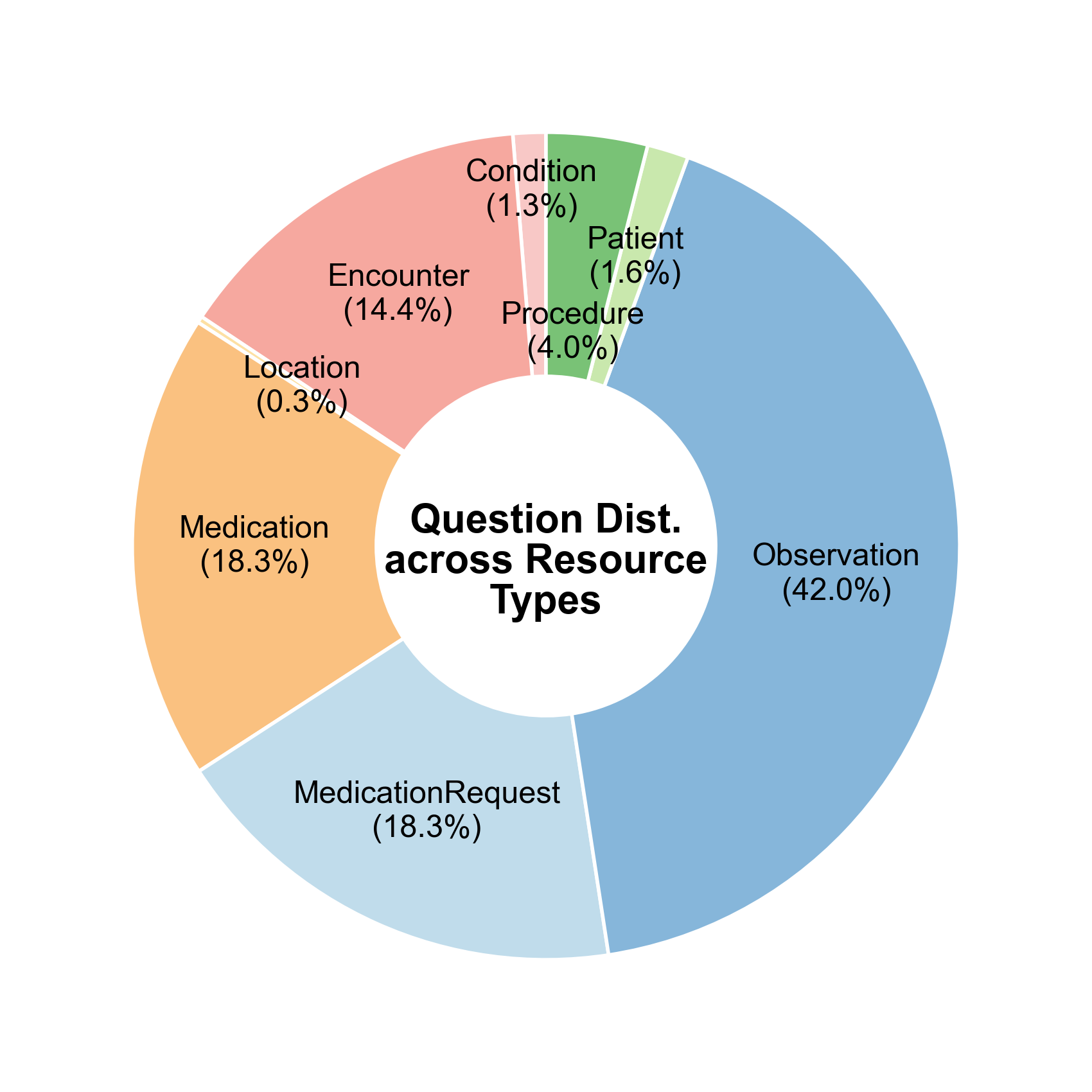}
\vspace{-15mm}
\caption{Distribution of questions by the FHIR resource type required to answer them. Observation includes input and output events, lab events, chart events, and microbiology, which constitutes the largest portion.}
\label{fig:resource_distribution}
\end{figure}

The final FHIR-AgentBench comprises 2,931 question–context–FHIR resource–answer pairs, grounded in real patient data from MIMIC-IV-FHIR (Table~\ref{tab:sample_questions}). Our benchmark covers a wide range of clinical topics in the questions, including observations, medications, and encounters (Figure~\ref{fig:resource_distribution}). Furthermore, FHIR-AgentBench contains a wide distribution of task complexity, as defined by the number of ground-truth FHIR ID resources required to answer each question. The number of ground-truth resources ranges from 0 to 2,257, with 80\% of the questions involving 0–2 resources and the remaining 20\% averaging around 70. This large variation poses two core challenges for the agent: (1) the agent cannot infer this number from the surface form of the question before interacting with the FHIR server, as even the same question with different patient IDs can yield vastly different results; and (2) given that the average number of FHIR resources per patient is about 9K, identifying the relevant ones is akin to finding a needle in a haystack.

These characteristics of FHIR-AgentBench reflect realistic and practical challenges for interoperable EHR question answering, such as reasoning temporally, navigating thousands of records, handling case-sensitive terminologies, and addressing queries with no relevant results.
\vspace{-2mm}

\vspace{-2mm}
\section{Experiments}
\label{sec:experiment}

To establish a performance baseline on FHIR-AgentBench, we conduct a comprehensive evaluation of LLM-based agents. Our experiments have three primary goals: (1) quantify the difficulty of the benchmark for common LLM architectures, (2) systematically diagnose key failure modes by ablating different agent design choices, and (3) understand the interplay between data retrieval accuracy and final answer correctness.

\subsection{Evaluation Framework}
We define the task as a two-stage problem: \textbf{Retrieval} and \textbf{Answer Generation}. For each question $q_i$, an agent must first retrieve a set of relevant FHIR resources $R_i$ and then use them to generate a final answer $\hat{y}_i$. The agent's outputs $\{(\hat{y}_i, \hat{R}_i)\}_{i=1}^N$ are evaluated against the ground-truth resources $R_i$ and answers $y_i$ using three standard metrics:
\begin{itemize}
    \itemsep0em
    \item \textbf{Retrieval Precision}: $P = \frac{|\hat{R}_i \cap R_i|}{|\hat{R}_i|}$
    \item \textbf{Retrieval Recall}: $R = \frac{|\hat{R}_i \cap R_i|}{|R_i|}$
    \item \textbf{Answer Correctness}: $AC = \mathbf{1}(\hat{y_i} = y_i)$
\end{itemize}

For cases where there are no true resources but predicted resources exist, precision is set to 0, and the case is removed from recall calculations. Conversely, when true resources exist but no predicted resources are retrieved, recall is set to 0, and the case is removed from precision calculations. In instances where there are neither true nor predicted resources, both precision and recall are assigned a value of 1, recognizing that the agent correctly identified the absence of relevant resources. 

Regarding answer correctness, we employ an LLM evaluator (OpenAI's o4-mini) to determine answer correctness, as agents generate answers in free-text. Manual review of 500 samples confirmed a 97\% agreement rate with human judgment, indicating high reliability (\appendixref{apd:evaluator_correctness}).
Additionally, no notable imbalance was present between misclassified correct and incorrect answers, indicating that the evaluator does not systematically overestimate or underestimate performance.

\subsection{Agent Architectures}
\label{sec:agent_architectures}

A preliminary experiment revealed that a naive approach of providing the full patient record in-context consistently failed due to the verbose, nested structure of FHIR data. On average, the full patient record in our benchmark contains about 3 million tokens when represented as FHIR JSON, which exceeds even larger context windows. This finding highlights the need for a sophisticated agent with dedicated retrieval capabilities. We therefore design and evaluate a spectrum of agent architectures built from three core tool components:

\begin{enumerate}
    \item \textbf{FHIR Query Generator}: A tool that enables the agent to compose and execute structured FHIR API queries as query strings (e.g., Observation?patient =XXX\&code=YYY\&date=geZZZ). This approach offers maximum flexibility for fine-grained data filtering but requires the agent to have precise knowledge of the FHIR API specification and the underlying data schema.
    \item \textbf{Retriever}: A set of specialized tools designed for simplified data access, allowing the agent to retrieve FHIR resources by specifying either a resource type and a patient FHIR ID, or a resource ID. The tools fetch all resources of the specified type for the given patient (e.g., retrieving all Observation resources for Patient XXX), or a single resource if its ID is provided.
    \item \textbf{Code Interpreter}: A tool that allows the agent to generate and execute Python code, with access to retrieved resources for reasoning, aggregation, and computation.
\end{enumerate}

We also evaluate two interaction patterns for agent implementation:
\begin{itemize}
    \item \textbf{Single-Turn Agents}: The agent must retrieve all necessary information and generate the final answer in a single, non-interactive pass. We test three variants: (1) FHIR Query Generator only, (2) Retriever only, and (3) Retriever followed by Code Interpreter.
    \item \textbf{Multi-Turn Agents (ReAct-style)}: The agent can iteratively reason and act, allowing it to refine its search or perform multi-step analysis. We test two variants: (1) Retriever only, and (2) Retriever with a Code Interpreter.
\end{itemize}
This systematic design allows us to isolate the impact of interaction patterns (single vs. multi-turn) and reasoning capabilities (retrieval vs. code) on overall performance.

\vspace{-2mm}
\subsection{Implementation Details}
For a robust comparison, we select a range of leading models, including both closed-source (o4-mini, Gemini-2.5-Flash) and open-source (Qwen3-32B, Llama-3.3-70B) alternatives as agent models. To ensure fairness, all models are restricted to a 32k-token context window. For each agent architecture and model, we calculate the average Precision, Recall, and Answer Correctness across all 2,931 questions in the benchmark.

\vspace{-2mm}
\section{Results}
\label{sec:result}

\subsection{Main Findings: A Challenging Benchmark for State-of-the-Art Agents}
Table \ref{tab:results_agents} presents the performance of baseline agents across the three evaluation metrics for o4-mini. 
Our experiments reveal that FHIR-AgentBench presents a significant challenge for current agent architectures. The highest-performing agent—a multi-turn agent with a retriever and code interpreter—achieves an answer correctness of only 50.0\%, underscoring the difficulty of reasoning over realistic, interoperable clinical data, as further illustrated in \appendixref{apd:results_agents_fhir_by_resource_type}.

Our systematic evaluation reveals two key trends. First, multi-turn interaction is crucial for high retrieval recall. Agents that can iteratively refine search consistently achieve higher recall (71\%) than single-turn agents (58\%). Second, code generation is essential for parsing complex FHIR data. In both single-turn and multi-turn settings, adding a code interpreter dramatically improves final answer correctness. This shows that LLMs struggle to interpret the nested, graph-like structure of FHIR resources without the procedural logic of code. 

Notably, retrieval precision is consistently low across all architectures. As described in Section \ref{sec:agent_architectures}, this is an expected consequence of the Retriever's design, which prioritizes recall over precision, thereby introducing significant noise the agent must navigate.

\paragraph{LLM model comparison} 
While there is variation in performance among models, no single model dominates across all metrics (Table \ref{tab:results_models}). For the top-performing multi-turn architecture, all models achieve answer correctness within a relatively narrow range (44-50\%). This suggests that the agent's architecture and the inherent difficulty of the task are currently more significant bottlenecks than the choice of the base LLM. Complete evaluation results for all models and approaches can be found in \appendixref{apd:results_agents_full}.

\paragraph{Effect of retrieval on answer correctness}
FHIR-AgentBench, unlike prior benchmarks, allows researchers to analyze how retrieval quality impacts downstream answer correctness. To illustrate this, we test the effect of precision on answer correctness in \figureref{fig:precision_graph}. We find that for all agents, higher retrieval precision correlates with significant gains in answer correctness. This confirms a critical trade-off: while high recall is necessary to obtain the right information, flooding the agent's context with irrelevant FHIR resources directly impairs its ability to find and reason over the correct data. Thus, even with models with larger context windows which make retrieving all required resources more probable, the agent still requires the correct procedural logic (i.e. code generation) to navigate references and parse the complex FHIR structure. This finding is consistent with recent studies that show  that curating task-relevant information is more effective than loading all available data into the context window, and that irrelevant content can degrade performance~\citep{asai2024selfrag, liu-etal-2024-lost, Anthropic2025ContextEngineering}.  

\paragraph{Need for architectural improvements} Higher precision improves answer correctness across agent architectures, but only up to a certain point - even in the highest precision bucket, correctness is far from 100\%. Beyond this retrieval-driven performance gain, our results suggest that better architectural design is required to push the performance boundaries and overcome fundamental reasoning challenges.

\begin{figure}[htbp]
\floatconts
  {fig:precision_graph}
  {\caption{Relative gain in answer correctness across precision buckets for the o4-mini model under different agent frameworks. Gains are calculated after filtering to questions with perfect recall, isolating the effect of unnecessary resources on answer correctness.}}
  {\includegraphics[width=\linewidth]{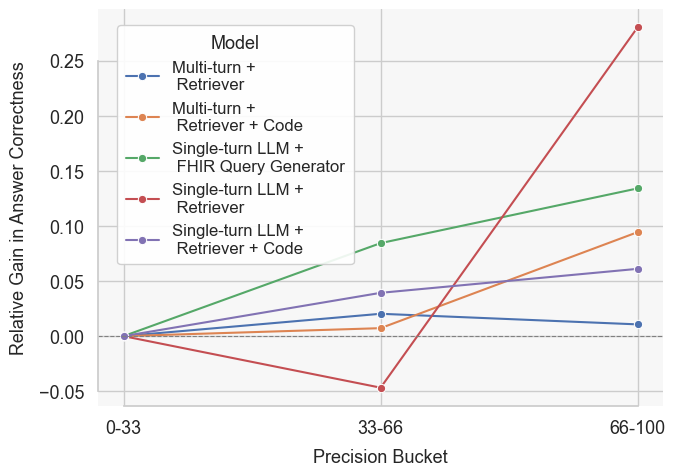}}
\end{figure}

\begin{table*}[h!]
\centering
\renewcommand{\arraystretch}{1.2}
\caption{Performance of different agentic approaches for \texttt{o4-mini} on FHIR-AgentBench.}
\label{tab:results_agents}
\small
\begin{tabular}{lccc}
\hline
\textbf{Agentic Approach} & \textbf{Precision} & \textbf{Recall} & \textbf{Answer Correctness} \\
\hline
Single turn LLM + FHIR Query Generator& 0.46& 0.43& 0.25\\
Single-turn LLM + Retriever& 0.41& 0.58& 0.22\\
Single-turn LLM + Retriever + Code& 0.41& 0.58& 0.33\\
Multi-turn + Retriever & 0.33& 0.71& 0.20\\
Multi-turn + Retriever + Code & 0.35 & 0.68& 0.50\\
\hline
\end{tabular}
\end{table*}

\begin{table*}[h!]
\centering
\renewcommand{\arraystretch}{1.2}
\caption{Performance comparison of different LLMs under the top agentic frameworks on FHIR-AgentBench.}
\vspace{-2mm}
\label{tab:results_models}
\small
\begin{tabular}{lccc}
\hline
\textbf{LLM} & \textbf{Precision} & \textbf{Recall} & \textbf{Answer Correctness} \\
\hline
\multicolumn{4}{l}{\textit{Single Turn LLM + Retriever + Code}} \\
\hspace{0.5cm} o4-mini & 0.41 & 0.58 & 0.33 \\
\hspace{0.5cm} Gemini-2.5-flash & 0.35 & 0.65 & 0.39 \\
\hspace{0.5cm} Qwen3 & 0.40 & 0.66 & 0.33 \\
\hspace{0.5cm} LLaMA-3.3 & 0.33 & 0.78 & 0.36 \\
\hline
\multicolumn{4}{l}{\textit{Multi-turn + Retriever + Code}} \\
\hspace{0.5cm} o4-mini & 0.35 & 0.68 & 0.50 \\
\hspace{0.5cm} Gemini-2.5-flash & 0.34 & 0.67 & 0.44 \\
\hspace{0.5cm} Qwen3 & 0.33 & 0.81 & 0.47 \\
\hspace{0.5cm} LLaMA-3.3 & 0.33 & 0.80 & 0.46 \\
\hline
\end{tabular}
\end{table*}

\subsection{Error Analysis}
To provide actionable insights for future research, we perform a detailed error analysis and display primary categories of failure.

\paragraph{Retrieval failures from misidentifying the search space}

Even with multi-turn capabilities, agents frequently fail at the initial step of identifying the correct FHIR resource type to search. For example, they often search for Procedures even though procedural outcomes are in Observations (see \appendixref{example1}). This highlights a fundamental challenge for agents in mapping clinical concepts to the specific, and often non-intuitive, structures of the FHIR schema.

Additionally, we observe tool-specific errors in frameworks that use the FHIR Query Generator as the retrieval mechanism, where the agent has full flexibility to compose FHIR queries. In this setup, the agent tends to generate \textbf{overly specific queries}, restricting the search too narrowly and leading to missed results, as in \appendixref{example2}. 

Lastly, in single-turn settings, the agent cannot retrieve \textbf{multiple resource types} required to answer the question. For instance, a query may require identifying the latest hospital encounter and checking for linked observations or procedures. However, since the agent is restricted to issuing only a single retrieval call with one resource type, it cannot subsequently retrieve the related resources, leading to incomplete reasoning and answers, such as in \appendixref{example3}. This highlights both the complexity of our question dataset and the need for multi-turn agents that can chain multiple retrieval steps.

\paragraph{Answer generation failures from failing to parse FHIR's complexity}

A significant source of error during answer generation, even when all correct resources are retrieved, is the agent's inability to correctly interpret the FHIR data structure, as illustrated in Appendix \ref{interpretation_examples}. This ``Interpretation Gap'' manifests in several ways:
\begin{itemize}
    \item \textbf{Navigating References:} For medication questions, agents consistently fail to follow the reference link from a MedicationRequest resource to its corresponding Medication resource, thus missing the actual medication name.
    \item \textbf{Incorrect Filtering:} For encounter-related questions, agents often count all Encounter resources instead of correctly filtering for inpatient visits, which may be nested within another Encounter resource, or fail to distinguish the current hospitalization from past ones.
    \item \textbf{Misinterpreting Fields:} For lab results, agents often filter on the human-readable text instead of the standardized FHIR code, leading to incorrect or incomplete results.
\end{itemize}

Lastly, the agent encounters \textbf{code execution failures}. Agents successfully retrieve data, but then generate erroneous Python code with syntax errors or incorrect logic. Without robust error handling, these failures prevent the agent from producing any answer, even with the correct information at hand.

\section{Conclusion}

This paper presents FHIR-AgentBench, a benchmark designed around three pillars of realism: authentic clinical questions, real patient data, and the mandated FHIR interoperability standard for evaluating LLM agents. With thousands of diverse clinical questions—from identifying diagnoses to calculating lab values—grounded in anonymized patient data, FHIR-AgentBench provides a realistic benchmark for the community. Notably, our inclusion of ground-truth FHIR resource mappings for each question enables a detailed assessment of an agent's retrieval and reasoning capabilities, moving beyond a single accuracy score to a precise diagnosis of model strengths and weaknesses.

Our comprehensive evaluation of common agent architectures reveals a critical bottleneck. While multi-turn architectures with code interpreters show the most promise, the top-performing agent achieves only 50\% accuracy. This demonstrates that the challenge lies not only in retrieving data but in the fundamental difficulty agents face in parsing the intricate, graph-like structure of FHIR data, even when the correct information is available.

By open-sourcing FHIR-AgentBench, we provide not only a dataset but also a research agenda for the clinical AI community.
Future work should focus on three key areas: improving the benchmark, the agents, and the evaluation methodologies.

First, we can expand our dataset to include more question categories, including multi-patient questions (i.e. ``How many patients were diagnosed with \{diagnosis\} this year?"). Currently, FHIR is inherently optimized for point-of-care scenarios (retrieving data for one patient), which results in retrieval latency dominating overall agent evaluation time for multi-patient retrieval. However, as hardware and FHIR APIs improve, we believe multi-patient questions will become efficiently answerable and relevant for population-level analysis. Additionally, we can also include unanswerable questions due to non-existent data sources (i.e. ``How much did this year's procedures cost for the patient?", for which FHIR-AgentBench does not have any cost data). These questions can further improve FHIR-AgentBench's clinical realism, where doctors may ask questions that summarize across patients, or questions for which no FHIR data exist. 

Second, we can improve the baseline agents' performance to determine how difficult FHIR-AgentBench is with more sophisticated frameworks. To improve retrieval, we can develop advanced retrieval tools that can retrieve multiple resources with finer filtering conditions. To improve answer generation, we can provide a more precise specification of the FHIR data structure through prompt tuning and targeted few-shot examples that synthesize nested information, particularly with medication questions. 

Lastly, while accuracy is crucial, future evaluations must also incorporate real-world constraints such as latency and cost, which are paramount for clinical deployment. 

FHIR-AgentBench provides the standardized, high-fidelity platform needed to diagnose failures, measure progress, and ultimately accelerate the development of safe, reliable, and truly interoperable AI for healthcare.

\acks{
This work was supported in part by the Institute of Information \& Communications Technology Planning \& Evaluation (IITP) grants (No.RS-2019-II190075), the Korea Health Industry Development Institute (KHIDI) grant (No.RS-2025-02213750), and National Research Foundation of Korea (NRF) grants (NRF-2020H1D3A2A03100945), funded by the Korea government (MSIT, MOHW).
}

\bibliography{jmlr-sample}

\clearpage

\appendix
\renewcommand\theHfigure{Appendix.\thefigure}
\renewcommand{\thefigure}{B\arabic{figure}}
\renewcommand\theHtable{Appendix.\thetable}
\renewcommand{\thetable}{A\arabic{table}} 

\setcounter{table}{0} 
\setcounter{figure}{0} 

\section{Supplementary Tables}

\subsection{Evaluator LLM Correctness}
Please refer to Table \ref{apd:evaluator_correctness}.

\begin{table*}[h!]
\centering

\caption{Robustness of LLM-based evaluation based on manual evaluation of 500 samples.}
\label{apd:evaluator_correctness}
\begin{adjustbox}{width=\linewidth,center}
\begin{tabular}{ccccc}
\hline
\makecell{\textbf{Correctness} \\ \textbf{Label}} &
\makecell{\textbf{\# Questions} \\ \textbf{Labeled by} \\ \textbf{Evaluator LLM}}  & 
\makecell{\textbf{\# Questions Match} \\ \textbf{between Human} \\ \textbf{and Evaluator LLM Labels}}  & 
\makecell{\textbf{Evaluator LLM} \\ \textbf{Accuracy}} & 
\makecell{\textbf{Evaluator LLM} \\ \textbf{Average Accuracy}} \\
\hline
Incorrect (0) & 169 & 160 & 0.95 & \multirow{2}{*}{0.97} \\
Correct (1) & 331 & 323 & 0.98 & \\
\hline
\end{tabular}
\end{adjustbox}
\end{table*}

\subsection{Agent Architecture Performance - by Model}

Please refer to Table \ref{apd:results_agents_full}.

\begin{table*}[h!]
\centering
\caption{Performance of different agentic approaches across multiple LLMs on FHIR-AgentBench.}
\small
\renewcommand{\arraystretch}{1.1}
\label{apd:results_agents_full}
\begin{tabular}{lcccc}
\hline
\textbf{Agentic Approach} & \textbf{Precision} & \textbf{Recall} & \textbf{Answer Correctness} & \\
\hline
\multicolumn{5}{l}{\textit{Single turn LLM + FHIR Query Generator}} \\
\hspace{0.5cm} o4-mini & 0.46 & 0.43 & 0.25\\
\hspace{0.5cm} Gemini-2.5-flash & 0.47 & 0.44 & 0.32\\
\hspace{0.5cm} Qwen3 &  0.51 & 0.43 & 0.30 \\
\hspace{0.5cm} LLaMA-3.3 & 0.58 & 0.40 & 0.27 \\
\hline
\multicolumn{5}{l}{\textit{Single Turn LLM + Retriever}} \\
\hspace{0.5cm} o4-mini & 0.41 & 0.58 & 0.22\\
\hspace{0.5cm} Gemini-2.5-flash & 0.41 & 0.59 & 0.21\\
\hspace{0.5cm} Qwen3 & 0.41 & 0.64 & 0.23\\
\hspace{0.5cm} LLaMA-3.3 & 0.34 & 0.78 & 0.17\\
\hline
\multicolumn{4}{l}{\textit{Single Turn LLM + Retriever + Code}} \\
\hspace{0.5cm} o4-mini& 0.41 & 0.58 & 0.33\\
 \hspace{0.5cm} Gemini-2.5-flash& 0.35 & 0.65 & 0.39\\
\hspace{0.5cm} Qwen3 & 0.40 & 0.66 & 0.33\\
\hspace{0.5cm} LLaMA-3.3 & 0.34 & 0.78& 0.36\\
\hline
\multicolumn{5}{l}{\textit{Multi-turn + Retriever}} \\
\hspace{0.5cm} o4-mini & 0.33 & 0.71 & 0.20 \\
\hspace{0.5cm} Gemini-2.5-flash & 0.34 & 0.70 & 0.20\\
\hspace{0.5cm} Qwen3 & 0.34 & 0.79 & 0.20\\
\hspace{0.5cm} LLaMA-3.3 & 0.33 & 0.80 & 0.16\\
\hline
\multicolumn{4}{l}{\textit{Multi-turn + Retriever + Code}} \\
\hspace{0.5cm} o4-mini& 0.35 & 0.68  & 0.50\\
 \hspace{0.5cm} Gemini-2.5-flash& 0.34 & 0.67 & 0.44\\
\hspace{0.5cm} Qwen3 & 0.33 & 0.81 & 0.47\\
\hspace{0.5cm} LLaMA-3.3 & 0.33 & 0.80 & 0.46\\
\hline
\end{tabular}
\end{table*}

\subsection{Agent Architecture Performance - by FHIR Resource Type}

Please refer to Table \ref{apd:results_agents_fhir_by_resource_type}.

\begin{table*}[h!]
\centering
\renewcommand{\arraystretch}{0.95}
\caption{Performance of different agentic approaches on FHIR-AgentBench by correct resource type, using o4-mini. NA values for precision occur when, for the given FHIR resource group, all questions have no predicted resources, and thus precision cannot be computed. We observe non-zero answer correctness when the recall is 0 because the agent still provides a final answer, which can be correct by chance, especially for binary Yes/No questions.}
\label{apd:results_agents_fhir_by_resource_type}
\small
\begin{tabular}{lccc}
\hline
\textbf{Agentic Approach / FHIR Resource} & \textbf{Precision} & \textbf{Recall} & \textbf{Answer Correctness} \\
\hline

\multicolumn{4}{l}{\textit{Single-turn LLM + Retriever}} \\
\hspace{0.5cm} Overall & 0.4102 & 0.5786 & 0.2177 \\
\hspace{0.5cm} Condition & 0.8499 & 0.4000 & 0.4286 \\
\hspace{0.5cm} Encounter & 0.3087 & 0.9949 & 0.8801 \\
\hspace{0.5cm} Location & NA & 0.0000 & 0.0000 \\
\hspace{0.5cm} Medication/MedicationRequest & 0.1644 & 0.0000 & 0.0502 \\
\hspace{0.5cm} Observation & 0.0333 & 0.4110 & 0.0201 \\
\hspace{0.5cm} Patient & 0.0100 & 1.0000 & 0.0000 \\
\hspace{0.5cm} Procedure & 0.1292 & 0.7037 & 0.6574 \\
\hspace{0.5cm} Empty & 1.0000 & 1.0000 & 0.2268 \\
\hline

\multicolumn{4}{l}{\textit{Single-turn LLM + FHIR Query Generator}} \\
\hspace{0.5cm} Overall & 0.4613& 0.4313& 0.2521 \\
\hspace{0.5cm} Condition & 0.8414 & 0.3429 & 0.3714 \\
\hspace{0.5cm} Encounter & 0.3039 & 0.9668 & 0.8214 \\
\hspace{0.5cm} Location & 0.3333 & 0.1429 & 0.1429 \\
\hspace{0.5cm} Medication/MedicationRequest & 0.1662 & 0.1429 & 0.0462 \\
\hspace{0.5cm} Observation & 0.0538 & 0.0855 & 0.0183 \\
\hspace{0.5cm} Patient & NA & 0.0000 & 0.0000 \\
\hspace{0.5cm} Procedure & 0.1293 & 0.6759 & 0.6667 \\
\hspace{0.5cm} Empty & 1.0000 & 1.0000 & 0.4094 \\
\hline

\multicolumn{4}{l}{\textit{Single-turn LLM + Retriever + Code}} \\
\hspace{0.5cm} Overall & 0.4082 & 0.5821 & 0.3333 \\
\hspace{0.5cm} Condition & 0.8499 & 0.4000 & 0.4286 \\
\hspace{0.5cm} Encounter & 0.3081 & 0.9974 & 0.5485 \\
\hspace{0.5cm} Location & NA & 0.0000 & 0.0000 \\
\hspace{0.5cm} Medication/MedicationRequest & 0.1634 & 0.0000 & 0.1004 \\
\hspace{0.5cm} Observation & 0.0329 & 0.4188 & 0.1972 \\
\hspace{0.5cm} Patient & 0.0100 & 1.0000 & 0.5682 \\
\hspace{0.5cm} Procedure & 0.1292 & 0.7037 & 0.5278 \\
\hspace{0.5cm} Empty & 1.0000 & 1.0000 & 0.5549 \\
\hline

\multicolumn{4}{l}{\textit{Multi-turn + Retriever}} \\
\hspace{0.5cm} Overall & 0.3342 & 0.7083 & 0.2003 \\
\hspace{0.5cm} Condition & 0.8398 & 1.0000 & 0.9714 \\
\hspace{0.5cm} Encounter & 0.3081 & 0.9974 & 0.8903 \\
\hspace{0.5cm} Location & 0.0323 & 1.0000 & 1.0000 \\
\hspace{0.5cm} Medication/MedicationRequest & 0.1291 & 0.1867 & 0.0502 \\
\hspace{0.5cm} Observation & 0.0228 & 0.6108 & 0.0070 \\
\hspace{0.5cm} Patient & 0.0100 & 1.0000 & 0.0000 \\
\hspace{0.5cm} Procedure & 0.1169 & 0.9722 & 0.9167 \\
\hspace{0.5cm} Empty & 1.0000 & 1.0000 & 0.0927 \\
\hline

\multicolumn{4}{l}{\textit{Multi-turn + Retriever + Code }} \\
\hspace{0.5cm} Overall & 0.3509 & 0.6830 & 0.5004 \\
\hspace{0.5cm} Condition & 0.8398 & 1.0000 & 1.0000 \\
\hspace{0.5cm} Encounter & 0.3043 & 0.9847 & 0.7933 \\
\hspace{0.5cm} Location & 0.0323 & 0.8571 & 0.8571 \\
\hspace{0.5cm} Medication/MedicationRequest & 0.1332 & 0.1988 & 0.1391 \\
\hspace{0.5cm} Observation & 0.0241 & 0.6367 & 0.3684 \\
\hspace{0.5cm} Patient & 0.0100 & 1.0000 & 0.5682 \\
\hspace{0.5cm} Procedure & 0.1148 & 0.9907 & 0.8056 \\
\hspace{0.5cm} Empty & 1.0000& 1.0000 & 0.7368 \\
\hline

\end{tabular}
\end{table*}

\section{Supplementary Figures}
\subsection{SQL-to-FHIR Translation}

\begin{figure*}[t!]
\centering
\vspace{5mm}
\includegraphics[width=\linewidth]{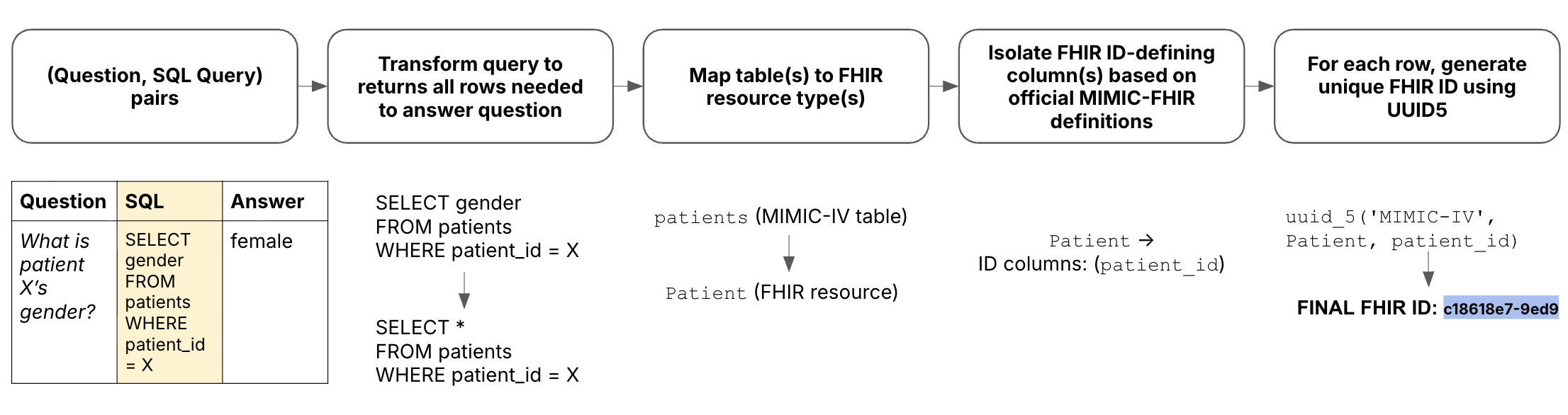}
\caption{Illustration of FHIR resource ID translation process from SQL query.}
\label{fig:translation_example}
\end{figure*}

Please refer to Figure \ref{fig:translation_example}.

\section{Error Examples} \label{error_ex}
\subsection{Retrieval Error Examples}
\subsubsection{Example 1} \label{example1}
Question : ``What is the number of times that patient XXX had ostomy (output) since 12/03/2133?''\\
True answer : 68\\
Agent answer : 0\\
Agent steps :
\begin{enumerate}
    \item The agent incorrectly queried \textbf{Procedure} resources (98 items retrieved).
    \item It wrote Python code to filter procedures for the display \textit{``Ostomy (output)''} between 2133-12-03 and 2133-12-31.
    \item No matches were found, so it answered: \textit{``The final answer is: 0.''}
\end{enumerate}
\textbf{Error:} Ostomy output measurements are stored in \textbf{Observation} resources (code \verb|226582|, category ``Output''), not in Procedure. Because the wrong resource type was queried, the correct answer (68) was missed.

\subsubsection{Example 2} \label{example2}
Question : ``When was the first time the respiratory rate of patient XXX was measured to be less than 23.0 today?''\\  
True answer : 2133-12-31T02:00:00-05:00\\  
Agent answer : No result found\\
Agent steps :
\begin{enumerate}
    \item The agent generated the query: 
    \begin{quote}
    \raggedright\texttt{Observation?patient=XXX}\\
    \texttt{\&code=9279-1}\\
    \texttt{\&date=ge2133-12-31T00:00:00}\\
    \texttt{\&date=le2133-12-31T23:59:59}
    \end{quote}
    \item The query returned no matching resources.
    \item The agent concluded: \textit{``No such observation was found today.''}
\end{enumerate}
\textbf{Error:} The agent restricted the query to the LOINC code \verb|9279-1| for respiratory rate. In this dataset, respiratory rate is instead coded as \verb|220210| (system: \texttt{chartevents-d-items}, display ``Respiratory Rate''). Because of this mismatch, the valid observation (value = 21 insp/min at 02:00) was missed.

\subsubsection{Example 3} \label{example3}
Question : ``Did patient XXX undergo an ultrasonography of superior vena cava, guidance treatment in the first hospital visit?''\\  
True answer : Yes\\  
Agent answer : Cannot determine / No\\  
Agent steps :
\begin{enumerate}
    \item The agent queried Encounters with: \verb|Encounter?patient=XXX|.
    \item It correctly retrieved the patient's first hospital encounter.
    \item However, the agent was restricted to a single retrieval call and could not subsequently query the linked Procedure resources.
    \item As a result, it concluded: \textit{``I cannot find the procedure during the first hospital encounter.''}
\end{enumerate}
\textbf{Error:} Answering the question required combining information from both \textbf{Encounter} and \textbf{Procedure} resources. Because the agent was limited to a single retrieval call, it failed to join these resource types and missed the correct answer.

\subsection{Interpretation Error Example}
\label{interpretation_examples}
\subsubsection{Example 1}
Question: ``Compute the number of times patient XXX was prescribed famotidine in their last hospital visit.''\\
True answer: 1\\
Agent answer: 0\\
Agent steps:
\begin{enumerate}
    \item The agent correctly retrieved \textbf{Encounter} and \textbf{MedicationRequest} resources (4 and 77 items, respectively).
    \item It wrote Python code to count MedicationRequest entries with \textit{medicationCodeableConcept} indicating ``famotidine'' in the last hospital encounter.
    \item Since all retrieved MedicationRequest entries had the \textit{medicationReference} field to refer to actual Medications, instead of \textit{medicationCodeableConcept}, no matches were found.
    \item It answered: \textit{``The final answer is: 0.''}
\end{enumerate}
\textbf{Error:} Famotidine prescriptions were recorded via \textbf{medicationReference} pointing to separate \textbf{Medication} resources, not inline in \textit{medicationCodeableConcept}. Because the agent only checked \textit{medicationCodeableConcept}, it missed the actual prescription (1).

\subsubsection{Example 2} 

Question: ``How many times patient XXX came to the hospital since 1 year ago?''\\
True answer: 1\\
Agent answer: 2\\
Agent steps:
\begin{enumerate}
    \item The agent retrieved all \textbf{Encounter} resources for the patient (2 items).
    \item It wrote Python code to count encounters with \textit{period.start} within the last year.
    \item The code did not distinguish hospital (inpatient) visits from ambulatory or other encounter types, so both encounters were counted.
    \item It answered: \textit{``The final answer is: 2.''}
\end{enumerate}
\textbf{Error:} One of the counted encounters was an ambulatory visit. Because the agent did not filter encounters for hospital/inpatient visits (e.g., via \textit{identifier.system} containing ``encounter-hosp'' or \textit{class.code} == 'IMP'), it overcounted hospital visits.

\subsubsection{Example 3}

Question: ``Count the number of times that patient XXX received a laboratory test for hemoglobin since 07/2184.''\\
True answer: 2\\
Agent answer: 3\\
Agent steps:
\begin{enumerate}
    \item The agent retrieved all \textbf{Observation} resources for the patient (176 items).
    \item It normalized \texttt{coding.display} and \texttt{code.text} to exactly ``hemoglobin.''
    \item It counted every match between 07/01/2184 and 12/31/2184.
    \item It included an additional Observation that contained the word ``Hemoglobin'' (e.g., part of a panel) but was not the specific lab-item code (51222) for the hemoglobin test of interest.
    \item The agent answered: \textit{``The final answer is: 3.''}
\end{enumerate}
\textbf{Error:} The agent matched on display text rather than the standard code identifier, causing one false positive. Only Observations with \texttt{code.coding} \texttt{system = `http://mimic. mit.edu/fhir/mimic/CodeSystem/}
\texttt{mimic-d-lab items'} and \texttt{code = `51222'} should be counted.

\section{Baseline Agent Prompts}
Please refer to Listing \ref{lst:single-querygenerator-prompt}, Listing \ref{lst:single-retriever-prompt}, Listing \ref{lst:single-retriever-code-prompt}, Listing \ref{lst:multi-retriever-prompt}, and Listing \ref{lst:multi-retriever-code-prompt} for the prompts used for each baseline agents.

\lstdefinestyle{mystyle}{
    backgroundcolor=\color{gray!15},   
    commentstyle=\color{codegreen},
    keywordstyle=\color{magenta},
    numberstyle=\tiny\color{codegray},
    stringstyle=\color{codepurple},
    basicstyle=\ttfamily\footnotesize,
    breakatwhitespace=false,         
    breaklines=true,            
    breakindent=0pt,
    captionpos=b,                    
    keepspaces=true,                 
    numbers=none,               
    numbersep=5pt,                  
    showspaces=false,                
    showstringspaces=false,
    showtabs=false,                  
    tabsize=2,
}
\lstset{style=mystyle}

\begin{figure*}[t]
\begin{lstlisting}[caption=Prompt for Single-turn + FHIR Query Generator Agent., label={lst:single-querygenerator-prompt}]
You are a helpful assistant that answers questions about patient data.

Use <FHIR QUERY GENERATOR TOOL> to retrieve patient data and answer questions based on the retrieved data.
Available resource types: <FHIR RESOURCE TYPES>. You can only call on these FHIR resources types for retrieval.

First, you should retrieve the patient data using <FHIR QUERY GENERATOR TOOL>. Then, you should reason about the retrieved data and provide the answer.
When you provide answers, make sure to provide them in the same format as they are in the retrieved data. If multiple answers are provided, provide them all in a list.
If you cannot find the answer or relevant patient data, clearly state that you cannot find the information.
Do not guess attributes; instead, use the provided tool to retrieve the data.
Do not get stuck or repeat the same action.
Do not plan ahead. Instead, directly use the tool to retrieve and reason about the data.

Few-shot examples:
<FEW SHOT EXAMPLES>

Follow these examples to structure your approach: retrieve relevant FHIR resources first, then analyze the data systematically using natural language reasoning.
\end{lstlisting}
\end{figure*}

\begin{figure*}[t]
\begin{lstlisting}[caption=Prompt for Single-turn + Retriever Agent., label={lst:single-retriever-prompt}]
You are a helpful assistant that answers questions about patient data.

Use <RETRIEVAL TOOLS> to retrieve patient data and answer questions based on the retrieved data. The available resource types are: <FHIR RESOURCE TYPES>. You can only call on these FHIR resources types for retrieval. 

First, you should retrieve the patient data using <RETRIEVAL TOOLS>. Then, you should reason about the retrieved data and provide the answer. When you provide answers, make sure to provide them in the same format as they are in the retrieved data. If multiple answers are provided, provide them all in a list. If you cannot find the answer or relevant patient data, clearly state that you cannot find the information.

Do not guess attributes; instead, use the provided tool to retrieve the data.
Do not get stuck or repeat the same action.
Do not plan ahead. Instead, directly use the tool to retrieve and reason about the data.

Few shot examples:
<FEW SHOT EXAMPLES>

Follow these examples to structure your approach: retrieve relevant FHIR resources first, then analyze the data systematically using natural language reasoning.

\end{lstlisting}
\end{figure*}

\begin{figure*}[t]
\begin{lstlisting}[caption=Prompt for Single-turn + Retriever + Code Agent., label={lst:single-retriever-code-prompt}]
You are a helpful assistant that answers questions about patient data.

You are given the following tools to retrieve and analyze patient data: <RETRIEVAL TOOLS>, <CODE GENERATION TOOL>.
The available resource types are: <FHIR RESOURCE TYPES>. You can only call on these FHIR resources types for retrieval.
First, you should retrieve the patient data using <RETRIEVAL TOOLS>. Then, you should reason about the retrieved data using <CODE GENERATION TOOL> and provide the answer based on the available information.

When you retrieve FHIR resources, they will be automatically available in your code execution environment as:
- retrieved_resources: dictionary with resource types as keys containing the retrieved FHIR data
    
When you provide answers, make sure to provide them in the same format as they are in the retrieved data. If multiple answers are provided, provide them all in a list.
If you cannot find the answer or relevant patient data, clearly state that you cannot find the information.
Do not guess attributes; instead, use the provided tool to retrieve the data.
Do not get stuck or repeat the same action.
Do not plan ahead. Instead, directly use the tool to retrieve and reason about the data.

Ensure code is syntactically correct and runnable without errors.
When generating code, make sure to first remove timezones so you can compare dates in a timezone-naive way without error.

Few-shot examples:
<FEW SHOT EXAMPLES>

Follow these examples to structure your approach: retrieve relevant FHIR resources first, then use Python code to analyze and process the data systematically.

\end{lstlisting}
\end{figure*}

\begin{figure*}[t]
\begin{lstlisting}[caption=Prompt for Multi-turn + Retriever Agent., label={lst:multi-retriever-prompt}]
You are a helpful assistant that can answer questions about patient data.

You have access to the following tools via function calling:
<RETRIEVAL TOOLS>

Available FHIR resource types: <FHIR RESOURCE TYPES>. You can only call on these FHIR resources types for retrieval.

To answer questions about patient data:
1. Use <RETRIEVAL TOOLS> to retrieve relevant FHIR resources
2. Analyze the retrieved data to answer the question
3. When you have completed your analysis and are ready to provide the final answer, you MUST format your response as follows:

   The final answer is: [your answer here]

If there are multiple answers, provide all of them.
IMPORTANT: Always end your response with 'The final answer is:' followed by your conclusion. This is required for proper processing.
When you provide answers, make sure to provide them in the same format as they are in the retrieved data. If multiple answers are provided, provide them all in a list.
If you cannot find the answer or relevant patient data, clearly state that you cannot find the information.
Do not guess attributes; instead, use the provided tool to retrieve the data.
Do not get stuck or repeat the same action.

Few-shot examples:
<FEW SHOT EXAMPLES>

Follow these examples to structure your approach: retrieve relevant FHIR resources first, then analyze the data systematically using natural language reasoning.


\end{lstlisting}
\end{figure*}

\begin{figure*}[t]
\begin{lstlisting}[caption=Prompt for Multi-turn + Retriever + Code Agent., label={lst:multi-retriever-code-prompt}]
You are a helpful assistant that can answer questions about patient data.

You have access to the following tools via function calling: <RETRIEVAL TOOLS>, <CODE GENERATION TOOL>

Available FHIR resource types: <FHIR RESOURCE TYPES>. You can only call on these FHIR resources types for retrieval.

To answer questions about patient data:
1. Use <RETRIEVAL TOOLS> to retrieve relevant FHIR resources
2. Use <CODE GENERATION TOOLS> to analyze the retrieved data
3. When you have completed your analysis and are ready to provide the final answer, you MUST format your response as follows:

   The final answer is: [your answer here]

When you retrieve FHIR resources, they will be automatically available in your code execution environment as:
- retrieved_resources: dictionary with resource types as keys containing the retrieved FHIR data

When calling <CODE GENERATION TOOL> and generating code, refer to the following pointers:
    - Ensure code is syntactically correct and runnable without errors.
    - Compare dates in a timezone-naive way.
If there are multiple answers, provide all of them.

IMPORTANT: Always end your response with 'The final answer is:' followed by your conclusion. This is required for proper processing.

When you provide answers, make sure to provide them in the same format as they are in the retrieved data. If multiple answers are provided, provide them all in a list. If you cannot find the answer or relevant patient data, clearly state that you cannot find the information.

Do not guess attributes; instead, use the provided tool to retrieve the data.
Do not get stuck or repeat the same action.

Few-shot examples:
<FEW SHOT EXAMPLES>

Follow these examples to structure your approach: retrieve relevant FHIR resources first, then use Python code to analyze and process the data systematically.


\end{lstlisting}
\end{figure*}

\end{document}